\documentclass[article,twoside,web]{ieeecolor}
\usepackage{tmi}
\usepackage{cite}
\usepackage{amsmath,amssymb,amsfonts}
\usepackage{graphicx}
\usepackage{textcomp}
\usepackage{multirow}
\usepackage{stfloats}
\usepackage{float} 
\usepackage{url}
\usepackage{verbatim}
\usepackage{pifont}
\usepackage{algorithmic}
\newcommand{\ie}{\textit{i}.\textit{e}.}
\newcommand{\eg}{\textit{e}.\textit{g}.}
\usepackage[colorlinks,hyperindex,breaklinks]{hyperref}
\newcommand{\myparagraph}[1]{\vspace{0.1em}\noindent\textbf{#1}}
\usepackage[table]{xcolor}
\usepackage{arydshln}
\definecolor{ours}{gray}{.95}

\def\BibTeX{{\rm B\kern-.05em{\sc i\kern-.025em b}\kern-.08em
    T\kern-.1667em\lower.7ex\hbox{E}\kern-.125emX}}
\begin{document}
\title{Merging Context Clustering with Visual State Space Models for Medical Image Segmentation}
\author{Yun Zhu,
Dong Zhang,~\IEEEmembership{Member,~IEEE},
Yi Lin, Yifei Feng,
Jinhui Tang,~\IEEEmembership{Senior Member,~IEEE}
\thanks{This work was supported by the Major Science and Technology Projects in Jiangsu Province under Grant BG2024042.} 
\thanks{Y. Zhu and J. Tang are with the School of Computer Science and Engineering, Nanjing University of Science and Technology, Nanjing 210094, China. E-mail: \{zhuyunmissy, jinhuitang\}@njust.edu.cn.}
\thanks{D. Zhang is with the Department of Electronic and Computer Engineering, The Hong Kong University of Science and Technology, Hong Kong, China. E-mail:~dongz@ust.hk.}
\thanks{Y. Lin is with the Department of Computer Science and Engineering, The Hong Kong University of Science and Technology, Hong Kong, China. E-mail:~yi.lin@connect.ust.hk.}
\thanks{Y. Feng is with the First School of Clinical Medicine, Nanjing Medical University; Department of General Surgery, the First Affiliated Hospital of Nanjing Medical University, Nanjing, China. E-mail:~fengyifei1982@163.com.}
\thanks{Corresponding authors: Jinhui Tang (jinhuitang@njust.edu.cn), Dong Zhang (dongz@ust.hk), and Yifei Feng (fengyifei1982@163.com).}
}
\maketitle
\begin{abstract}
Medical image segmentation demands the aggregation of global and local feature representations, posing a challenge for current methodologies in handling both long-range and short-range feature interactions. Recently, vision mamba (ViM) models have emerged as promising solutions for addressing model complexities by excelling in long-range feature iterations with linear complexity. However, existing ViM approaches overlook the importance of preserving short-range local dependencies by directly flattening spatial tokens and are constrained by fixed scanning patterns that limit the capture of dynamic spatial context information. To address these challenges, we introduce a simple yet effective method named context clustering ViM (CCViM), which incorporates a context clustering module within the existing ViM models to segment image tokens into distinct windows for adaptable local clustering. Our method effectively combines long-range and short-range feature interactions, thereby enhancing spatial contextual representations for medical image segmentation tasks. Extensive experimental evaluations on diverse public datasets, \ie, Kumar, CPM17, ISIC17, ISIC18, and Synapse demonstrate the superior performance of our method compared to current state-of-the-art methods. Our code can be found at \href{https://github.com/zymissy/CCViM}{https://github.com/zymissy/CCViM}.
\end{abstract}
\begin{IEEEkeywords}
Context clustering, Medical image segmentation, Vision mamba, Visual state space model.
\end{IEEEkeywords}
\section{Introduction}
\label{introduction}
\IEEEPARstart{M}{edical} image segmentation (MedISeg) plays a crucial role in the communities of scientific research and medical care, aiming at delineating the region of interest and recognizing pixel-level and fine-grained lesion objects~\cite{wang2021annotation,litjens2017survey,wang2022boundary,zou2022graph}, which is critical for anatomy research, disease diagnosis, and treatment planning~\cite{bai2020population,mei2020artificial}. Deep learning advancements have sped up the automation of MedISeg, resulting in improved efficiency, accuracy, and reliability when compared to manual segmentation technologies~\cite{clinically,fully}. In the past years, accurate MedISeg methods have been successfully used in daily clinical applications, \eg, nuclei segmentation in microscopy images~\cite{graham2019hover,sun2023automated}, skin lesion segmentation in dermoscopy images~\cite{esteva2017dermatologist}, and multi-organ segmentation in CT images~\cite{gibson2018automatic,qi2024exploring}. 

\begin{figure*}[t]
\centerline{\includegraphics[width=.95\textwidth]{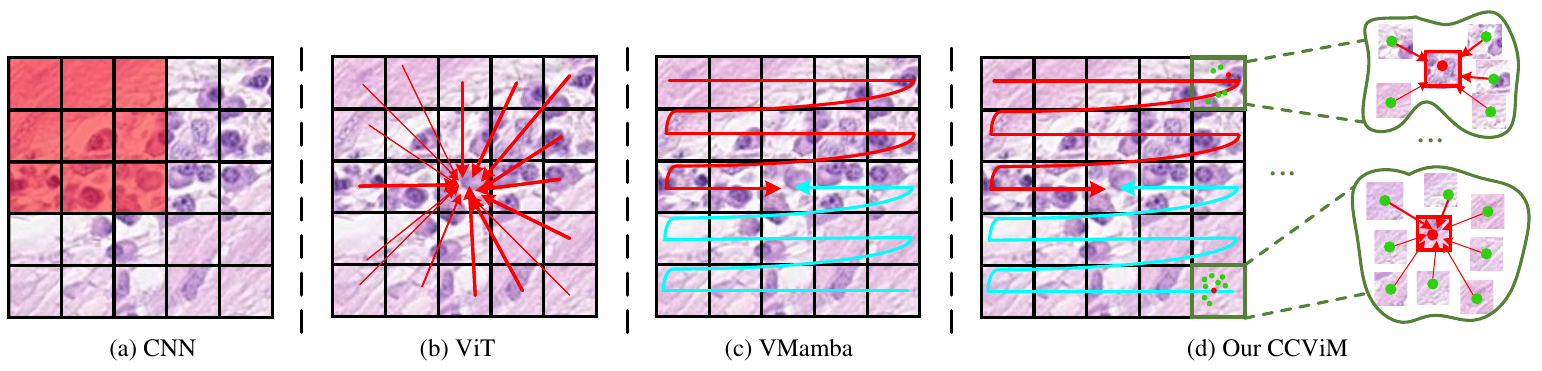}}
\vspace{-3mm}
\caption{Illustration of the feature interaction mechanism in MedISeg. CNNs~\cite{deepcnn} in (a) conceptualize an image as structured feature grids, employing convolutional layers that slides over the local space with a certain stride. ViT~\cite{vit} in (b) uses self-attention, treating an image as tokens, enabling each token to interact with other tokens. VMamba~\cite{vmamba} in (c) uses the cross-scan module to integrate the pixels from different directions, achieving a global receptive field with linear complexity. Our CCViM in (d) combines the cross-scan module with our context clustering layer. Our method treats an image as a set of data points, dynamically grouping all points into clusters within a local window to extract local contexts.}
\label{fig1}
\vspace{-4mm}
\end{figure*}

Various types of feature interactions in deep learning methods have a significant impact on the performance and capabilities of MedISeg models~\cite{zhang2020feature,yan2023progressive,yan2020higcin}. Fig.~\ref{fig1} illustrates different forms of feature interaction mechanisms in various MedISeg models. Convolutional neural networks (CNNs)~\cite{deepcnn,internimagecnn,Mobilenets} in Fig.~\ref{fig1} (a) conceptualize an image as a grid of pixels and compute element-wise multiplication between kernel weights and pixel values in a local field. While CNNs are effective at capturing local features, their ability to model long-range feature interactions is limited, which affects their performance. To address this limitation, Vision Transformer (ViTs) models have been introduced~\cite{transunet,zhang2023augmented,conformer}. As illustrated in Fig.~\ref{fig1} (b), ViTs treat an image as a sequence of patches, similar to tokens in natural language processing (NLP), allowing each token to attend to every token in the image and enabling ViTs to model long-range feature interactions. Due to the long-range feature interactions of ViT models, they can capture global contexts and demonstrate superior performance in MedISeg tasks~\cite{transunet,Att-UNet,Swin-unet}. However, the quadratic complexity of ViTs' long-range feature interactions in relation to the number of tokens has posed challenges in applying downstream applications.

To address these challenges, an efficient visual state space model called VMamba is introduced with linear complexity~\cite{vmamba,localmamba}. VMamba draws inspiration from the Mamba, originally designed for NLP tasks~\cite{Mamba}. The Mamba model processes input data causally, leveraging its causal properties to perform effectively and efficiently in language modeling tasks. However, the non-causal nature of images poses challenges for the Mamba model in handling image data using its causal processing approach. In contrast, VMamba addresses this issue by patching and flattening an image, as illustrated in Fig.~\ref{fig1} (c), then scanning the flattened image to integrate information from all locations in various directions, creating a global receptive field without increasing linear computational complexity. This innovative approach has led to the development of several Mamba-based models in the MedISeg domain~\cite{Vm-unet,HC-Mamba}. Despite the benefits of global scanning methods, they may overlook local features and spatial context information. LocalMamba introduces a local scanning method that confines the scan within a local window, enabling the capture of local dependencies while maintaining a global perspective~\cite{localmamba}. However, existing fixed scanning approaches may struggle to adaptively capture spatial relationships.

To address the above problems, we introduce a simple yet effective Context Clustering Vision Mamba (CCViM) model, which is a U-shaped architecture designed for MedISeg that effectively captures both global and local feature interactions. As illustrated in Fig.~\ref{fig1} (d), our CCViM can address the limitations of fixed scanning strategies by proposing the context clustering selective state space model (CCS6) layer, which combines the cross-scan layer with our context clustering (CC) layer to capture local spatial visual contexts dynamically. Compared to convolution scanning images in a grid-like manner or attention exploring mutual relationships within a sequence, our CC layer views the image as a set of data points and organizes these points into different clusters, which can dynamically capture local feature interactions. Comprehensive experiments are conducted on nuclei segmentation~\cite{kumar2017dataset,cpm}, skin lesion segmentation~\cite{isic2017,isic2018}, and multi-organ segmentation~\cite{Synapse} tasks to demonstrate the superior performance effectively and efficiently of our CCViM in MedISeg.

The main contributions of this paper are as follows: \emph{(1)} {Introduction of a novel Mamba-based model} for MedISeg, CCViM, which is effective and efficient. 
\emph{(2)} {Proposal of the CCS6 layer}, which combines global scanning directions with the CC layer, enhancing the model's feature representation capability.
\emph{(3)} {Proposal of the CC layer}, which can dynamically capture spatial contextual information, improving model performance compared to fixed scanning strategies.
\emph{(4)} {Comprehensive experiments} conducted on five public MedISeg datasets show that CCViM outperforms existing models, demonstrating superior performance.
\section{RELATED WORK}
\subsection{Medical Image Segmentation (MedISeg)}
\label{II-A}
MedISeg is crucial for physicians in diagnosing specific diseases, as it delineates the regions of interest with precision~\cite{wang2021annotation,litjens2017survey,wang2022boundary}. Automatic MedISeg has been extensively researched to address the limitations of manual segmentation, which is often time-consuming, labor-intensive, and subjective~\cite{bai2020population,zhang2022unabridged,mei2020artificial}. Consequently, deep learning methods such as convolutional neural networks (CNNs)~\cite{deepcnn,internimagecnn,Mobilenets} and vision transformers (ViTs)~\cite{attention,vit,Swin-unet,transunet} have been widely applied in MedISeg tasks~\cite{zhang2022understanding}. For instance, UNet~\cite{unet} is favored for its simple architecture and robust scalability, which derives many U-shaped models for MedISeg tasks~\cite{zhang2022unabridged}. UNet++~\cite{unet++} proposes nested encoder-decoder sub-networks with skip connections for MedISeg. TransUnet~\cite{transunet} proposes the first Transformer-based model for MedISeg tasks. Swin-Unet~\cite{Swin-unet}, a pure Transformer-based U-shaped model, adopts a hierarchical Swin Transformer~\cite{swintransformer} block with shifted windows as the encoder and performs effectively in MedISeg. However, CNNs' performance is limited with the local respective field, and Transformer-based models require quadratic complexity. To overcome these challenges, Mamba~\cite{Mamba} has been proposed to solve the computational efficiency problem and modeling long-range dependencies. In this paper, our research is also based on Mamba. Our contribution is to propose a more efficient local interaction mechanism.

\subsection{Feature Interaction for MedISeg}
\label{II-B}
Feature interaction matters, and it greatly affects MedISeg's performance. CNNs have dominated the field of computer vision in recent years, benefiting from some key inductive biases \eg, locality and translation equivalence~\cite{wang2022convolutional,zhang2020feature}. The CNN's locality helps identify distinct and stable points in images, \eg, corners, edges, and textures, which are significant for detecting small and irregularly shaped target features in MedISeg~\cite{shi2024focusdet,olson1997automatic}. However, the local feature interaction cannot accurately address the complex and interconnected anatomical structures in MedISeg, which also requires long-range feature interactions to capture the global context and spatial relationships within the medical image~\cite{zhang2020causal}. In recent years, ViTs have emerged as effective long-range feature interaction methods in MedISeg tasks, \eg, Swin-Unet~\cite{Swin-unet}, UTNet~\cite{gao2021utnet}, and Missformer~\cite{huang2022missformer}. Since ViTs abandon the local bias from CNNs, some researchers combine convolution and attention to acquire both local and global features, \eg, TransUnet~\cite{transunet} and Conformer~\cite{conformer}. Although the above methods inherit the advantages from short-range and long-range features and achieve better performance, the insights and knowledge are still restricted to CNNs and ViTs \cite{cocs}. Besides, CNNs or ViTs are limited in modeling long-range dependencies or computational efficiency. Therefore, except for Mamba~\cite{Mamba}, we also attempt to employ a new paradigm for visual representation by using a clustering~\cite{cocs}. A clustering algorithm views an image as a set of points, allowing it to capture local topology information adaptively without introducing significant computational complexity. 
\begin{figure*}[t]
\centerline{\includegraphics[width=1\textwidth]{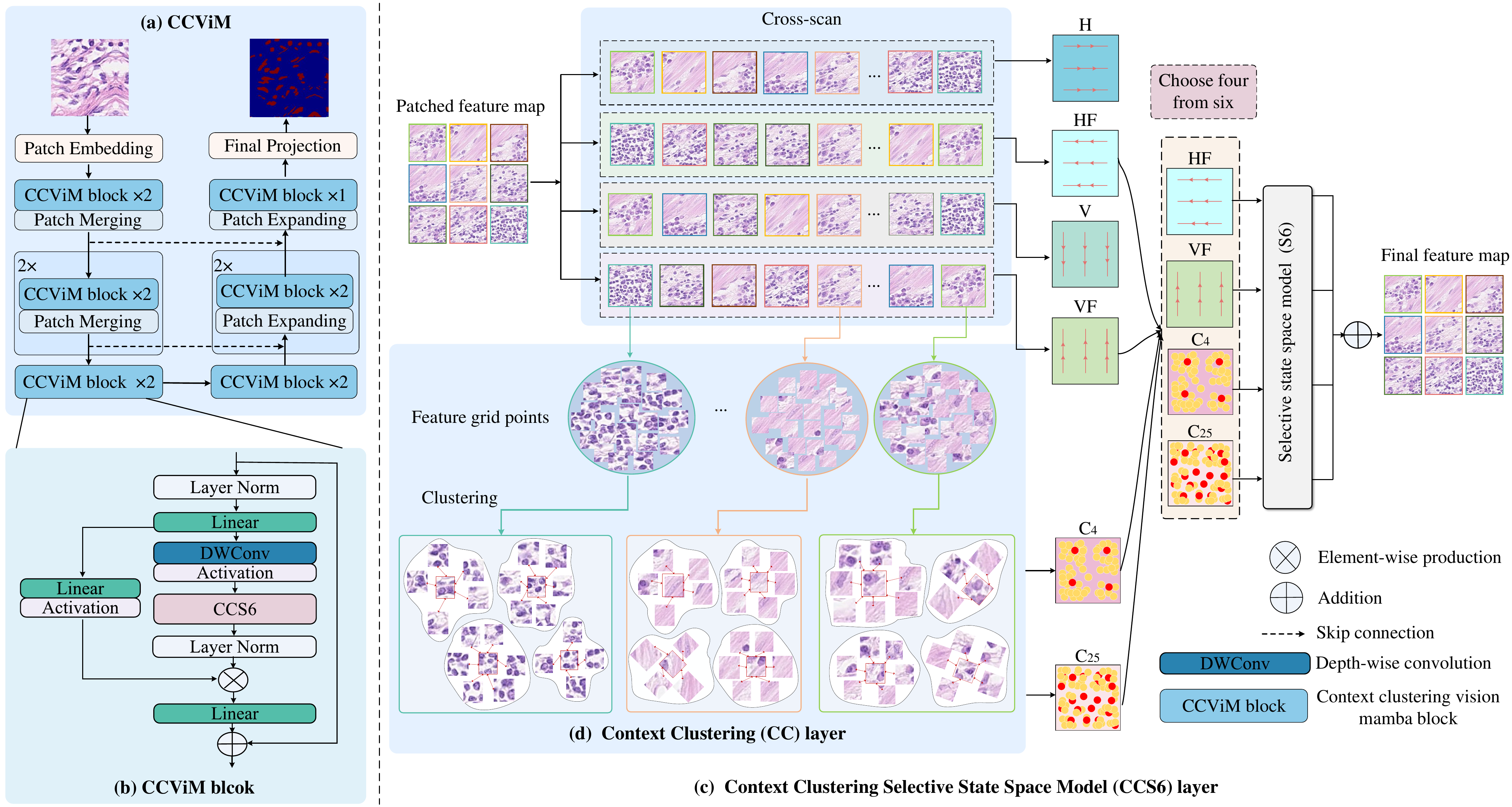}}
\vspace{-2mm}
\caption{(a) The overall architecture of context clustering vision mamba (CCViM), a U-shaped structure for MedISeg. (b) CCViM block, the core component of CCViM. (c) CCS6 layer, the core module of CCViM block. CCS6 layer employs six different methods to process the input (patched feature maps). Four of these methods use a cross-scan module to flatten the patched feature maps and scan the flattened features in four different directions. The other two methods apply CC layers with $4$ and $25$ cluster centers. Then select four methods from all six methods to process the input feature map, and input the processed information into the S6 module. Finally, merge the output features to construct the final feature map. (d) CC layer, which views each patch as a set of feature grid points, and clusters these feature grid points into several centers.}
\label{fig2}
\vspace{-3mm}
\end{figure*}

\subsection{Vision Mamba (ViM)}
\label{II-C}
Mamba \cite{Mamba}, originally designed for the NLP tasks, proposes a novel selective mechanism that focuses on relevant information or filters out irrelevant information. Mamba~\cite{Mamba} eliminates the limitations in CNNs and Transformers, and achieves both effectiveness and efficiency. However, due to the intrinsic non-causal characteristic of images, there is a significant obstacle to Mamba's comprehension of images. Therefore, a series of Mamba-based models have been proposed to address this issue. For instance, the representative ViM~\cite{vim} ﬂattens spatial data into 1D tokens and scans these tokens in two different directions to address the non-casual problem in the vision domain. However, flattening spatial data into 1D tokens disrupts the natural 2D dependencies. To address this problem, VMamba~\cite{vmamba} introduces a cross-scan module to bridge the gap between 1D array scanning and 2D dependencies, effectively resolving the non-causal problem in visual images without compromising the receptive field. Besides, influenced by the success of VMamba \cite{vmamba}, VM-UNet \cite{Vm-unet} introduces a visual Mamba UNet for MedISeg. However, these Mamba-based methods overlook the local features, which are both important in the vision domain. Therefore, LocalMamba~\cite{localmamba} introduces a local scan strategy to capture both global and local information. However, such fixed scanning approaches are limited in their ability to dynamically capture spatial relationships. To address this challenge, we propose the CCS6 module, which integrates scanning modules with our CC to adaptively extract both global and local features while capturing spatial context information.

\section{Our Method}
\subsection{Preliminary and Notation}
\label{III-A}
Mamba~\cite{Mamba} is a causal visual state space model, which is both effective and efficient, identifying that a key weakness of the structured state space model (SSM) is its inability to perform content-based reasoning. To overcome this, Mamba introduces a selective state-space model (S6). The recurrent formula of SSM can be defined as follows:
\begin{equation}
\label{eq1}
\begin{aligned}
h^{\prime}(t) &= \overline{\boldsymbol{A}}h(t) + \overline{\boldsymbol{B}}x(t),\\
y(t) &= \boldsymbol{C}h(t),
\end{aligned}
\end{equation}
where $h(t)\in\mathbb{R}^{N}$ is the intermediate latent state to map $x(t)\in\mathbb{R}$ to $y(t)\in\mathbb{R}$. $\boldsymbol{A}\in\mathbb{R}^{N\times N}$ and $\boldsymbol{B}\in \mathbb{R}^{N\times N}$ are discretized by $\overline{\boldsymbol{A}} = \exp(\Delta \boldsymbol{A})$ and $\overline{\boldsymbol{B}} = (\Delta \boldsymbol{A})^{-1}(\exp(\Delta\boldsymbol{A})-\boldsymbol{E}) \cdot \Delta\boldsymbol{B}$. $\Delta$ is the timescale parameter for discretizing the parameters $\boldsymbol{A}$ and $\boldsymbol{B}$. Through discretization, continuous-time system SSM can be integrated into deep-learning models. $\boldsymbol{C}\in\mathbb{R}^{N \times 1}$ is the parameter matrix. $\boldsymbol{E}$ is the identify matrix. On the basis of SSM, S6 lets the SSM parameters be functions of the input, allowing the model to selectively focus on or filter out information. Making the parameters $\Delta, \boldsymbol{B}, \boldsymbol{C} $ depend on the input $x$, and the formulas are defined as follows:
\begin{equation}
\label{eq2}
\begin{aligned}
    s_{B}(x) &= \mathrm{Linear}_{N}(x),\\
    s_{C}(x) &= \mathrm{Linear}_{N}(x),\\
    s_{\Delta}(x) &= \mathrm{softplus}(\mathrm{Parameter}+\\ &\mathrm{Broadcast}_{D}(\mathrm{Linear}_{1}(x))),
\end{aligned}
\end{equation}
where $\mathrm{Linear}_{N}, \mathrm{Linear}_{1}$ are fully connected (FC) layers to project the embedding dimension of $x$ to $N$ and $1$. $\mathrm{Broadcasrt}_{D}$ matches the dimension of the output of $x$ to the Parameter. softplus is an activation function.
Integrating S6, the Mamba not only achieves linear computation, but also performs excellent in language modeling tasks~\cite{Mamba,parkcan,glorioso2024zamba,lieber2024jamba}. However, due to the inherent non-causal characteristics of images, it’s difficult for Mamba-based models~\cite{Mamba} to handle image data well. Fortunately, VMamba~\cite{vmamba} proposes a cross-scan module to scan the patched and flattened image in four directions, which can address the non-causality of the image without compromising the field of reception and increasing computational complexity.

\subsection{Overall Architecture}
\label{III-B}
The overview of CCViM is illustrated in Fig.~\ref{fig2} (a), which is composed of a patch embedding layer, an encoder, a decoder, a final projection layer, and skip connections. CCViM is not a symmetrical structure like UNet~\cite{unet}, but adopts an asymmetric design. First, input the image $\boldsymbol{I}\in\mathbb{R}^{3\times W\times H}$ into the patch embedding layer to divide $\boldsymbol{I}$ into non-overlapping patches of size $4\times 4$, getting a feature map with dimensions of $C\times \frac{W}{4}\times \frac{H}{4}$, where $C$ is the number of feature channels. Then, input the feature map into the encoder network. 
Each stage in the encoder network is composed of two CCViM blocks and one patch merging layer, while the last stage only has two CCViM blocks without patch merging layer after them. The patch merging layer is for reducing the height and width of the feature maps and increasing the number of channels, and the channels of feature maps in the four stages are $[C, 2C, 4C, 8C]$. Thirdly, the feature maps are translated into the decoder, which also has four stages. Each stage of decoder is composed of one patch expanding layer and two CCViM blocks, while the first stage only has two CCViM blocks without patch expanding layer before them. The patch expanding layer is for increasing the height and width of the feature maps and reducing the number of channels, and the channels of feature maps in the four stages are $[8C, 4C, 2C, C]$. Furthermore, we simply employ the addition skip connections to capture low-level and high-level features. Besides, we adopt the most foundational Cross-Entropy and Dice loss for our MedISeg tasks.
The CCViM block is the core component of CCViM, as shown in Fig.~\ref{fig2} (b). The input is first translated into the normalization layer and then split into two branches. In the first branch, the input passes through a linear layer followed by an activation layer. In the second branch, the input passes through a linear layer, depth-wise convolution, and an activation layer and then translates into the core module of the CCViM block: context clustering selective state space model (CCS6) layer. After normalizing the output of CCS6 layer, multiply it with the output from the first branch to merge the out of the two pathways. Finally, a linear layer is used to project the merged features onto the dimensions of the initial input features to establish residual connections.

\subsection{Context Clustering Selective State Space Model}
\label{III-C}
Context clustering selective state space model (CCS6) layer is the core component of the CCViM block, which adopts a selective state space model (S6) as its footstone. S6 processes the input data causally, resulting in only capturing vital information within the scanned part of the data, which is difficult for processing non-causal images. 
Based on which, numerous researches have also proposed various scan strategies to solve this problem well~\cite{vim,vmamba,Vm-unet,localmamba}. However, these global scanning methods overlook the local features and the spatial context information in medical images. Furthermore, all these fixed scanning approaches cannot effectively capture spatial relationships adaptively.
To overcome these challenges, we propose a CCS6 layer to extract both global and local features while capturing spatial context information in a learnable way. 
We adopt VMamba's~\cite{vmamba} cross-scan module, which proposes a selective scan mechanism across four different directions. As shown in Fig.~\ref{fig2} (c), we patch and flatten the input image, and scan the flattened image in both horizontal and vertical directions to capture the global information. At the same time, our CC layer performs learnable local context clustering in local windows. Additionally, the CC layer employs two different numbers of clustering centers--4 and 25--to capture varying structural information. Consequently, there are two distinct CC layers and four different scanning directions available for selection. To avoid introducing too much computational complexity, in each CCS6 layer, we select four from the six choices as in~\cite{Vm-unet,localmamba}. The detailed configuration of these choices will be introduced in section{~\ref{IV-B}.
\subsection{Context Clustering}
\label{III-D}
Instead of using convolution or attention to extract information, we use the context cluster (CC) algorithm \cite{cocs} for local extraction and spatial context information. In our CC layer, we view the image as a set of data points and group all points into clusters. Instead of clustering data points over the entire image, which will bring a significant computational cost. We split the image into distinct windows and limit the clustering operation to a local region. In each cluster, we aggregate the points into a center adaptively. As shown in Fig.~\ref{fig2} (d), after the patch embedding layer, the image $\boldsymbol{I}\in \mathbb{R}^{3\times W\times H}$ is transformed into patched feature maps $\boldsymbol{F}\in \mathbb{R}^{d\times w\times h}$. We convert the patched feature maps into a set of data points $\boldsymbol{P}\in \mathbb{R}^{d\times n}$, where $n=w\times h$ represents the total number of data points, $d$ is the point feature dimension. These points are unordered and disorganized. We then group the data points into several clusters based on their similarities, ensuring that each point is assigned to only one cluster. 

For CC layer, We first project $\boldsymbol{P}$ to $\boldsymbol{P}_S\in\mathbb{R}^{n\times d'}$ to compute the similarity, where $d'$ is the new point feature dimension. We evenly propose $t$ centers in each local window and compute the center feature by averaging its $k$ nearest points. Then, we calculate the pair-wise cosine similarity between the resulting $t$ center points and $\boldsymbol{P}_S$ to get the similarity matrix $\boldsymbol{S}\in \mathbb{R}^{t\times n}$. Based on the similarity, we allocate each point to the most similar center to get $t$ clusters. Furthermore, each cluster may exhibit a varying number of data points. In exceptional cases, some clusters may contain no data points, indicating that they are redundant. During the clustering, all data points in one cluster are dynamically aggregated into the center point based on the similarities. In a cluster, there is a small set of $m$ data points, represented by $\boldsymbol{P}_m \in \mathbb{R}^{m\times d'}$. Calculating the similarity between the $m$ data points $\boldsymbol{P}_m \in \mathbb{R}^{m\times d'}$ and the center, which is represented by $\boldsymbol{s}\in \mathbb{R}^{m}$ and is the subset of similarity matrix $\boldsymbol{S}$. We map the $m$ data points $\boldsymbol{P}_m \in \mathbb{R}^{m\times d'}$ to a value space to get $\boldsymbol{P}'_{m}\in \mathbb{R}^{m\times d'}$. We also generate a value center $\boldsymbol{v}_{c}$ of the $\boldsymbol{P}'_{m}$ in the value space, which is like the clustering center proposal. Then the aggregated feature $\boldsymbol{g} \in \mathbb{R}^{d'}$ is formulated as follow:
\begin{equation}
\label{eq3}
\begin{aligned}
\boldsymbol{g} = \frac{1}{T}\left(\boldsymbol{v}_{c} +\sum_{i=1}^{m} \operatorname{\sigma}\left(\alpha \boldsymbol{s}_{i} + \beta\right)*\boldsymbol{p}_{i}\right),\\
\mathrm{s.t.},\quad T=1+\sum_{i=1}^{m}\operatorname{\sigma} \left(\alpha \boldsymbol{s}_{i}+\beta\right),
\end{aligned}
\end{equation}
where $\alpha$ and $\beta$ are learnable scalars  used to scale and shift the similarity, and $\sigma$ is a sigmoid function that re-scales the similarity to the range (0, 1). $\boldsymbol{p}_{i}$ denotes $i$-th point in $\boldsymbol{P}'_{m}$. For numerical stability and to emphasize locality, we incorporate the value center $\boldsymbol{v}_{c}$ in Eq.~\eqref{eq3}. To control the scale, normalizing the aggregated feature by a factor of $T$.
Subsequently, adaptively dispatching the aggregated feature $\boldsymbol{g}$ to each data point in a cluster based on the similarity. This approach facilitates the communication of the points in the cluster, enabling them to share features in the cluster. For each data point $\boldsymbol{p}_{i}$ of $\boldsymbol{P}'_{m}$ in the cluster, updating it by:
\begin{equation}
\label{eq4}
\begin{aligned}
\boldsymbol{p}_i' = \boldsymbol{p}_i + \mathrm{FC} (\operatorname{\sigma}(\alpha \boldsymbol{s}_{i} + \beta)*\boldsymbol{g}),
\end{aligned}
\end{equation}
where $\boldsymbol{s}$ denotes similarity, using the same procedures as above. We apply a fully connected (FC) layer to project the feature dimension from the value space dimension $d'$ to the original dimension $d$. \vspace{-2mm}

\subsection{Post Processing}
\label{III-E} 
In the nuclei segmentation task (\ie, Kumar~\cite{kumar2017dataset} and CPM17~\cite{cpm}), we employ the post-processing method watershed algorithm to distinguish between individual nuclei instances. Following research~\cite{graham2019hover}, we create horizontal and vertical distance maps by calculating the distances from nuclear pixels to their centers of mass in both the vertical and horizontal directions. Within our model we predict the vertical distance maps $P_v$ and horizontal distance maps $P_h$. Additionally, we apply the Sobel operator to these distance maps to obtain the horizontal and vertical gradient maps. We then select the maximum value between these gradient maps: $M_s = \max(\text{Sobel}(P_v), \text{Sobel}(P_h))$. This process aids in edge detection by calculating the gradient magnitude, thereby emphasizing areas of significant intensity change. Next, we calculate the markers, $M$, by applying a threshold function to the probability map $P$ and gradient map $M_s$, where the markers are defined as $M=\delta(\tau(P,r)-\tau(M_s,k))$. Here, $\tau(P,r)$ is a threshold function,  with $r$ being the threshold value. If the value exceeds $r$, it is set to 1; otherwise, it is set to 0. $\delta$ is used to set negative values to 0. We then obtain the energy landscape $E=(1-\tau(M_s, k))\times\tau(P, h)$. Finally, $M$ serves as the marker during marker-controlled watershed to determine how to split $\tau(P,h)$, guided by the energy landscape $E$.\vspace{-4mm}

\subsection{Loss Function}
\label{III-F}
For a fair comparison with other state-of-the-art methods, we employ the most fundamental loss functions in medical image segmentation: Cross-Entropy and Dice loss~\cite{Vm-unet,graham2019hover,V-net}. Cross-entropy ensures pixel-level classification accuracy, while Dice loss addresses the common issue of class imbalance by optimizing the overlap between the predicted and true segmentation~\cite{liu2024we}. As shown in Eq.~\eqref{eq5} and Eq.~\eqref{eq6}, we combine Cross-Entropy and Dice loss to balance precise pixel-wise classification with overall segmentation performance.
\begin{equation}
\begin{aligned}
\label{eq5}
L=L_{\mathrm{Ce}}+L_{\mathrm{Dice}}
\end{aligned}
\end{equation}
\begin{equation}
\begin{aligned}
\begin{cases}
\label{eq6}
&L_\text{Ce}=-\frac{1}{N}\sum_{i=1}^N\sum_{c=1}^Cy_{i,c}\log(\hat{y}_{i,c})\\
\\
&L_\text{Dice}=1-\frac{2\sum_{i=1}^Ny_{i}\hat{y}_{i}+1}{\sum_{i=1}^Ny_{i}+\sum_{i=1}^N\hat{y}_{i}+1}
\end{cases}
\end{aligned}
\end{equation}
where, $N$ denotes the total number of samples, and $C$ represents the total number of categories. $y_{i, c}$ is an indicator of ground truth, equals 1 if sample $i$ belongs to category $c$, and 0 if it does not. $\hat{y}_{i, c}$ is the probability that the model predicts sample $i$ as belonging to category $c$. $y_{i}$ and $\hat{y}_{i}$ represent the ground truth and prediction, respectively.

\section{Experiments}
\label{exper}
\subsection{Datasets and Evaluation Metrics}
\label{IV-A} 
\begin{figure}[htbp]
\centerline{\includegraphics[width=\columnwidth]{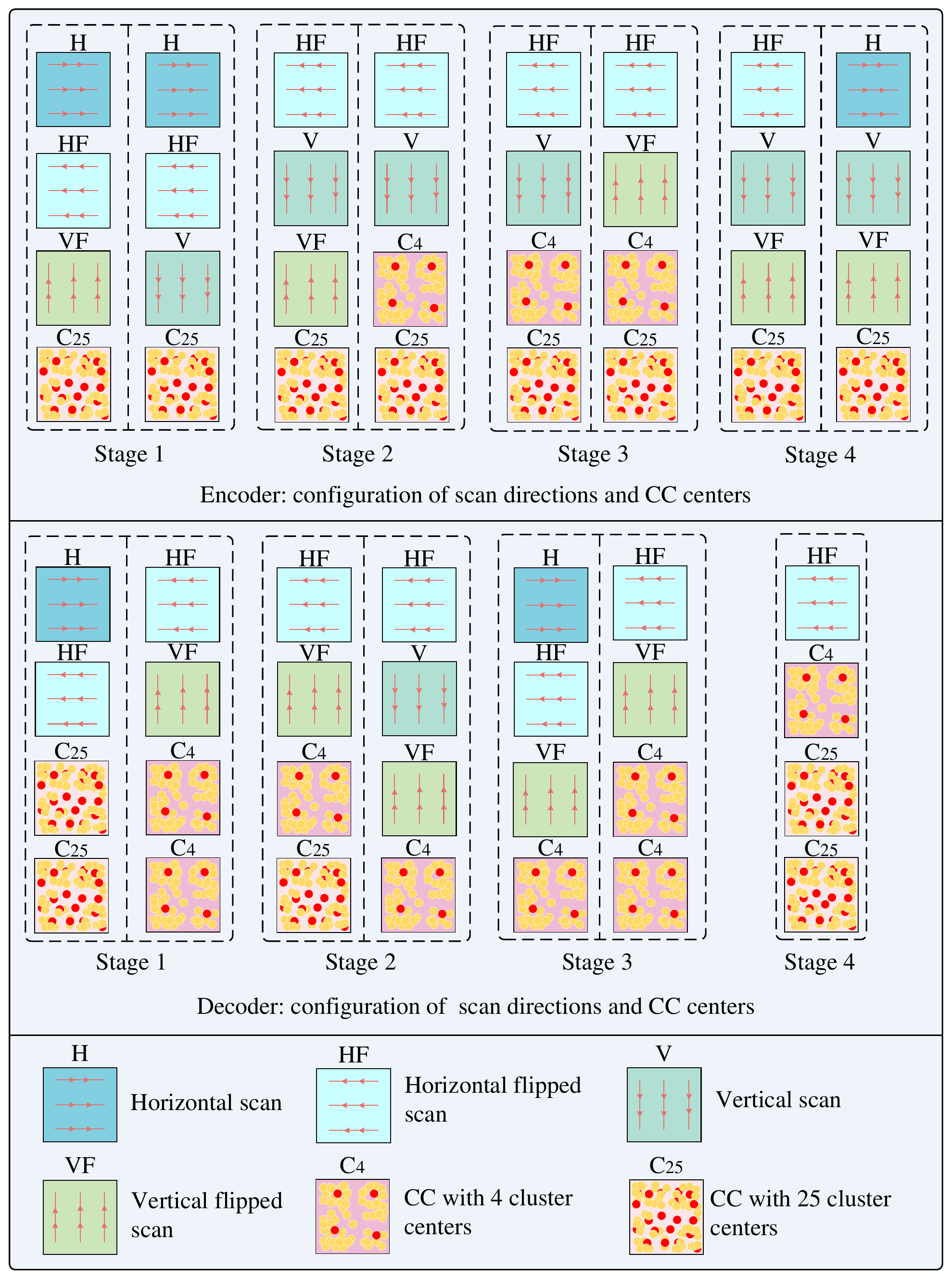}}
\vspace{-2mm}
\caption{The configuration of various scan directions and CC layer with different cluster centers. The configuration of each stage is different. We directly use the configuration of LocalMamba~\cite{localmamba}, replacing LocalMamba's local scan with our CC layer.}
\label{fig3}
\vspace{-4mm}
\end{figure}
\myparagraph{Datasets.} We evaluate CCViM on five MedISeg datasets, which contain nuclei segmentation datasets (\ie, Computational Precision Medicine (CPM17)~\cite{cpm} and Kumar~\cite{kumar2017dataset}), skin lesion segmentation datasets (\ie, ISIC17~\cite{isic2017} and ISIC18~\cite{isic2018}), and Synapse multi-organ segmentation dataset (\ie, Synapse~\cite{Synapse}). These datasets are detailed as follows:
\begin{itemize}
\item \textbf{Kumar:} The size of images is $1000\times 1000$ pixels at $40\times$ magnification. The total number of nuclei is 21623. Following~\cite{kumar2017dataset}, we split the dataset into two different sub-datasets: (i) Kumar-Train, a training set with 16 images, and (ii) Kumar-Test, a test set with 14 images. Following previous research~\cite{graham2019hover}, we crop each training image into $540 \times 540$ with an overlap of 164 pixels, then resize into $250 \times 250$. Data augmentation, including flip and rotation, is applied to all methods. In the inference, the images are cropped into $250 \times 250$ with an overlap of 164 pixels. 
\item \textbf{CPM17}: The size of given images is $500\times 500$ to $600 \times 600$ pixels at 40$\times$ or 20$\times$ magnification. The total number of nuclei is 7570. Following the challenge~\cite{cpm}, we split 32 images in the training dataset and 32 images in the test dataset. The data processing also follows the previous research\cite{graham2019hover}. 
\item \textbf{ISIC17 \& ISIC18:} The International Skin Imaging Collaboration 2017 and 2018 challenge datasets (ISIC17~\cite{isic2017} and ISIC18~\cite{isic2018}) contain 2,150 and 2,694 images with segmentation masks, respectively. Following previous research~\cite{MALUNet}, we allocate a 7:3 ratio for training and test sets, resize the images to $256\times 256$, and apply data augmentation like flip and rotation. 
\item \textbf{Synapse:} Synapse multi-organ segmentation dataset~\cite{Synapse} comprises 30 abdominal CT cases with 3779 axial abdominal clinical CT images, including 8 types of abdominal organs (aorta, gallbladder, left kidney, right kidney, liver, pancreas, spleen, stomach). Following previous research~\cite{transunet,Swin-unet}, we allocate 18 cases for training and 12 cases for testing. We resize the images to $224\times 224$ and apply augmentation including flip and rotation.
\end{itemize}

\myparagraph{Evaluation metrics.} (1) In nuclei segmentation~\cite{kumar2017dataset,cpm}, we employ the ensemble dice (DICE), aggregated Jaccard index (AJI), panoptic quality (PQ), detection quality (DQ), and segmentation quality (SQ) as the main evaluation metrics. PQ is composed of DQ and SQ, offering precise quantification and interpretability for evaluating the performance of nuclei segmentation. 
(2) In skin lesion segmentation, to compare our CCViM with previous methods on ISIC17~\cite{isic2017} and ISIC18\cite{isic2018} datasets, we employ mean intersection over union (mIoU), dice similarity coefficient (DSC), accuracy (Acc), sensitivity (Sen), and specificity (Spe) as the main evaluation metrics. 
(3) In Synapse~\cite{Synapse} multi-organ segmentation~\cite{Synapse}, we employ dice similarity coefffcient (DSC) and the $95\%$ Hausdorff distance (HD95) as the main evaluation metrics. 

\subsection{Configuration of Scan Directions and Local Clusters}
\label{IV-B}
In this paper, we do not apply the redundancy 22 scan strategies~\cite{rethinking} in each CCS6 layer, and there are only one, two, or three different scanning directions in each CCS6 layer. To extract local features and capture the spatial context information adaptively, we integrate our CC layer into each CCS6 layer. As shown in Fig.~\ref{fig3}, each CCS6 layer contains only four modules, which include one, two, or three different scanning directions and one or two different CC layers. This design avoids adding additional computational overhead compared to previous models. There are a total of four different scanning directions to choose from, including horizontal, horizontal flipped, vertical, and vertical flipped. Given the different sizes and number of image targets, we adopt two different CC layers. One CC layer method has 4 cluster centers in a local region, and the other one has 25 cluster centers in a local region. In nuclei segmentation tasks~\cite{kumar2017dataset,cpm}, there are many nuclei in one local region. In skin image segmentation tasks~\cite{isic2017,isic2018}, the target is large and may need lower point centers in one local region. We adopt the configuration provided by LocalMamba~\cite{localmamba}, with our CC lyaer replacing the local scan. Compared with the fixed local scan strategies in LocalMamba~\cite{localmamba}, our CC layer can dynamically cluster local features and capture spatial information adaptively.

\subsection{Implementation Details}
\label{IV-C}
In our all experiments, we set the batch size to 32 and employ AdamW~\cite{AdamW} optimizer with an initial learning rate of 1e-3. CosineAnnealingLR~\cite{scheduler} is employed as the scheduler. We set the training epochs to 300. All experiments initialize the models with ImageNet~\cite{imagenet} pretrained weights. All experiments are conducted on the PyTorch deep learning platform with a single NVIDIA GeForce RTX 3090 GPU.
\begin{figure}[htbp]
\centerline{\includegraphics[width=1\columnwidth]{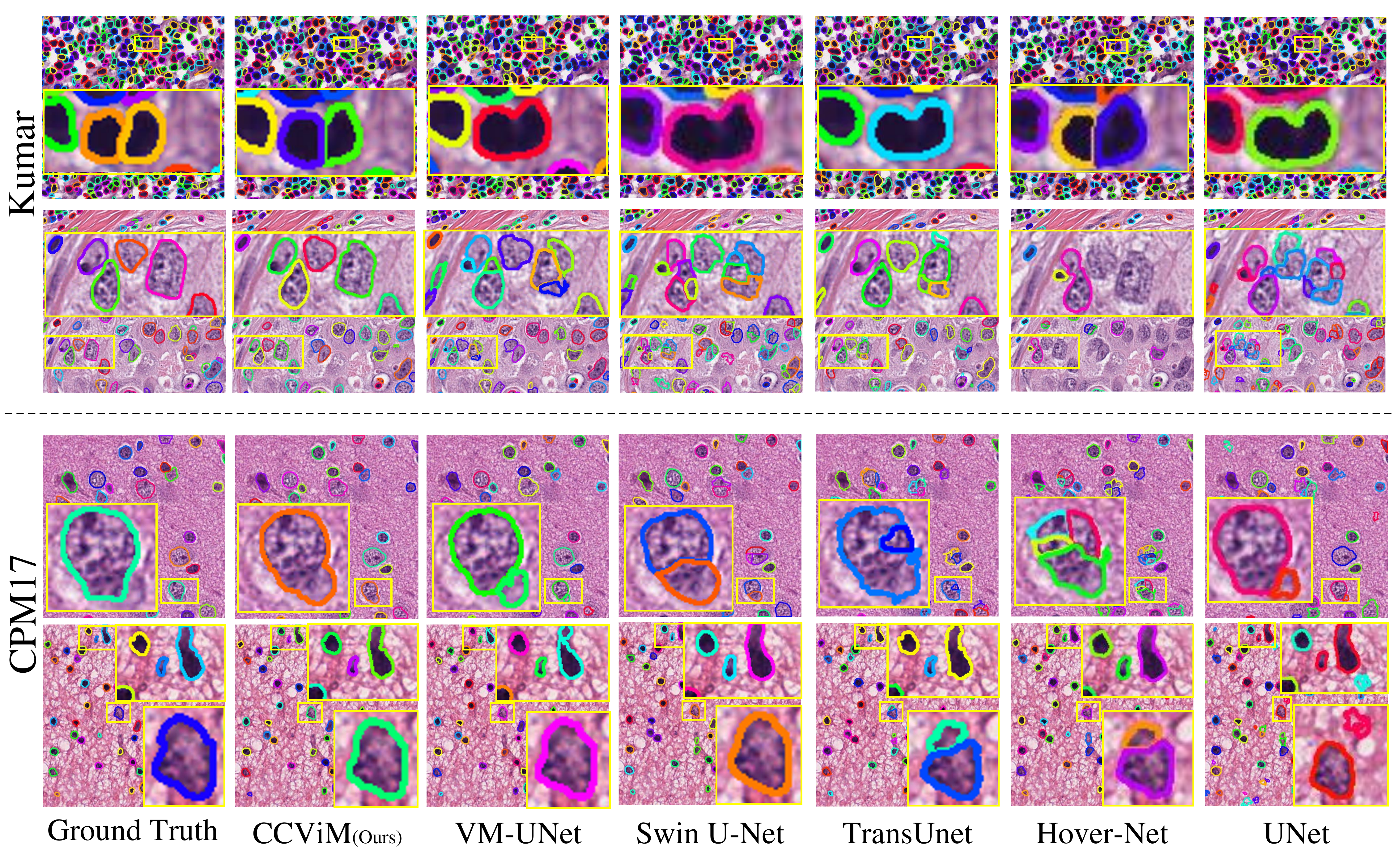}}
\vspace{-2mm}
\caption{Visualizations on Kumar~\cite{kumar2017dataset} and CPM17~\cite{cpm} datasets. Different colours of the nuclear boundaries denote separate instances. }
\label{fig4}
\vspace{-1mm}
\end{figure}
\begin{table}[t]
\centering
\renewcommand\arraystretch{1.2}
\setlength{\tabcolsep}{2.5pt}
{
\caption{Result comparisons with state-of-the-art methods on the Kumar~\cite{kumar2017dataset} and CPM17~\cite{cpm} datasets.}
\begin{tabular}{lr|ccccc} 
\hline 
~ & \textbf{Methods} & \textbf{PQ}(\%)\textuparrow & \textbf{Dice}(\%)\textuparrow & \textbf{AJI}(\%)\textuparrow & \textbf{DQ}(\%)\textuparrow & \textbf{SQ}(\%)\textuparrow \\			
\hline
\multirow{8}*{\rotatebox{90}{\textbf{Kumar}~\cite{kumar2017dataset}}} & SegNet\cite{segnet} & 40.70 & 81.10 & 37.70 & 54.50 & 74.20  \\
~ & UNet \cite{unet} & 47.80 & 75.80 & 55.60 & 69.10 & 69.00 \\ 
~ & DIST \cite{DIST} & 44.30 & 78.90 & 55.90 & 60.10 & 73.20 \\
~ & CIA-Net\cite{DIST} & 57.70 & 81.80 & \cellcolor[gray]{.95}\textbf{62.00} & 75.40 & 76.20 \\
~ & Hover-Net\cite{graham2019hover} & 58.22 & 81.32 & 59.62 & 75.62 & 76.71  \\
~ & TransUNet\cite{transunet} & 48.14 & 77.76 & 52.90 & 65.74 & 72.90 \\
~ & SwinUNet\cite{Swin-unet} & 47.46 & 80.11 & 51.01 & 64.68 & 72.86 \\
~ & VM-UNet \cite{Vm-unet} & 56.59 & 81.89	& 60.00 & 74.27 & 75.85 \\
~ & \cellcolor[gray]{.95}\textbf{CCViM$_{{\textrm{(Ours)}}}$} & \cellcolor[gray]{.95}\textbf{58.83} &
\cellcolor[gray]{.95}\textbf{82.48} & 61.38 & \cellcolor[gray]{.95}\textbf{76.50} & 
\cellcolor[gray]{.95}\textbf{76.63} \\
\hline
\multirow{8}*{\rotatebox{90}{\textbf{CPM17}~\cite{cpm}}} & SegNet\cite{segnet} & 53.10 & 85.70 & 49.10 & 67.90 & 77.80 \\
~ & UNet \cite{unet} & 57.80 & 81.30 & 64.30 & 77.80 & 73.40 \\
~ & DIST \cite{DIST} & 50.40 & 82.60 & 61.60 & 66.30 & 75.40 \\
~ & CIA-Net\cite{DIST} & 65.70 & 86.20 & 68.30 & 81.10 & 80.40 \\
~ & Hover-Net\cite{graham2019hover} & 70.47 & 87.63 & 72.44 & 86.67 & 81.14  \\ 
~ & TransUNet\cite{transunet} & 60.67 & 84.51 & 64.19 & 78.33 & 77.15 \\
~ & SwinUNet\cite{Swin-unet} & 60.12 & 86.08 & 62.88 & 77.11 &  77.52 \\
~ & VM-UNet \cite{Vm-unet} & 71.29 & 87.91 & 72.16 & 87.14 & \cellcolor[gray]{.95}\textbf{81.67} \\
~ & \cellcolor[gray]{.95}\textbf{CCViM$_{{\textrm{(Ours)}}}$} & \cellcolor[gray]{.95}\textbf{71.77} & \cellcolor[gray]{.95}\textbf{88.35} & \cellcolor[gray]{.95}\textbf{72.98} & \cellcolor[gray]{.95}\textbf{87.96} & 81.43 \\
\hline  
\end{tabular}
\label{tab1}}
\vspace{-1mm}
\end{table}
\begin{table}[t]
  \centering
  \renewcommand\arraystretch{1.2}
  \setlength{\tabcolsep}{2pt}
  {
   \caption{Result comparisons with state-of-the-art methods on the ISIC17~\cite{isic2017} and ISIC18~\cite{isic2018} datasets}
  \begin{tabular}{lr|ccccc} 
    \hline 
     ~ & \textbf{Methods}& \textbf{mIoU}(\%)\textuparrow & \textbf{DSC}(\%)\textuparrow & \textbf{Acc}(\%)\textuparrow & \textbf{Spe}(\%)\textuparrow & \textbf{Sen}(\%)\textuparrow 
     \\
     \hline
     \multirow{7}*{\rotatebox{90}{\textbf{ISIC17}~\cite{isic2017}}} & UNet\cite{unet} & 76.98 & 86.99 & 95.65 & 97.43 & 86.82 \\
     ~ & UTNetV2\cite{UTNetV2} & 77.35 & 87.23 & 95.84 & 98.05 & 84.85  \\
     ~ & TransFuse\cite{Transfuse} & 79.21 & 88.40 & 96.17 & 97.98 & 87.14  \\
     ~ & MALUNet\cite{MALUNet} & 78.78 & 88.13 & 96.18 & \cellcolor[gray]{.95}\textbf{98.47} & 84.78  \\
     ~ & VM-UNet \cite{Vm-unet} & 80.23 & 89.03 & 96.29 & 97.58 &
     \cellcolor[gray]{.95}\textbf{89.90}  \\
      ~ & HC-Mamba \cite{HC-Mamba} & 79.27 & 88.18 & 95.17 & 97.47 & 86.99  \\
     ~ & \cellcolor[gray]{.95}\textbf{CCViM$_{{\textrm{(Ours)}}}$}  & 
     \cellcolor[gray]{.95}\textbf{81.40}  &
     \cellcolor[gray]{.95}\textbf{89.74}  &
     \cellcolor[gray]{.95}\textbf{96.60}  & 98.19 & 88.70 \\
     \hline
     \multirow{10}*{\rotatebox{90}{\textbf{ISIC18}~\cite{isic2018}}}& UNet\cite{unet} & 77.86 & 87.55 & 94.05 & 96.69 & 85.86  \\
     ~ & UNet++\cite{unet++} & 78.31 & 87.83 & 94.02 & 95.75 & 88.65  \\
     ~ & Att-UNet\cite{Att-UNet} & 78.43 & 87.91 & 94.13 & 96.23 & 87.60 \\
     ~ & UTNetV2\cite{UTNetV2} & 78.97 & 88.25 & 94.32 & 96.48 & 87.60  \\
     ~ & SANet\cite{SANet} & 79.52 & 88.59 & 94.39 & 95.97 & 89.46  \\
     ~ & TransFuse\cite{Transfuse} & 80.63 & 89.27 & 94.66 & 95.74 & 
     \cellcolor[gray]{.95}\textbf{91.28}  \\
     ~ & MALUNet\cite{MALUNet} & 80.25 & 89.04 & 94.62 & 96.19 & 89.74  \\
     ~ & VM-UNet \cite{Vm-unet} & 81.35 & 89.71 & 94.91 & 96.13 & 91.12   \\
      ~ & HC-Mamba \cite{HC-Mamba} & 80.60 & 89.25 & 94.84 & 97.08 & 87.90  \\
     ~ & \cellcolor[gray]{.95}\textbf{CCViM$_{{\textrm{(Ours)}}}$} & \cellcolor[gray]{.95}\textbf{81.92} & \cellcolor[gray]{.95}\textbf{90.06} & \cellcolor[gray]{.95}\textbf{95.23} & \cellcolor[gray]{.95}\textbf{97.32} & 88.74   \\
     \hline  
   \end{tabular}
   \label{tab2}
  }
\vspace{-3mm}
\end{table}

\subsection{Comparisons with State-of-the-art Methods}
\label{IV-D}
\myparagraph{Results on nuclei segmentation.}
We compare our CCViM with CNN-based, Transformer-based, and Mamba-based models. Table~\ref{tab1} shows the results of different models on Kumar~\cite{kumar2017dataset} and CPM17~\cite{cpm} datasets. In Table~\ref{tab1}, the TransUnet~\cite{transunet} and Swin U-Net~\cite{Swin-unet} have superior results than U-Net~\cite{unet}, especially on CPM17~\cite{cpm} datasets. Compared to the methods above, VM-UNet~\cite{Vm-unet} demonstrates enhanced performance, particularly on the PQ and Dice metrics. This demonstrates the effectiveness of the Mamba-based model. On the Kumar dataset~\cite{kumar2017dataset}, VM-UNet~\cite{Vm-unet} performs worse than Hover-Net~\cite{graham2019hover} due to the small and dense nature of the nuclei on this dataset, which requires local feature interactions to capture the subtle details of these objects. In contrast, our CCViM outperforms Hover-Net~\cite{graham2019hover}, improving the PQ, Dice, AJI, DQ, and SQ by $0.61\%$, $1.16\%$, $1.76\%$, $0.88\%$ and $-0.08\%$, respectively. This demonstrates that our CCViM is superior at capturing local features. On the CPM17 dataset, compared to VM-UNet~\cite{Vm-unet}, our CCViM has improved the PQ, Dice, AJI, DQ, and SQ by $0.48\%$, $0.44\%$, $0.82\%$, $0.82\%$ and $-0.24\%$, respectively. Overall, our CCViM has the best performance in the PQ metric, indicating its ability to achieve more precise separation of individual nuclei. As shown in Fig.~\ref{fig4}, our CCViM demonstrates superior performance on both the Kumar~\cite{kumar2017dataset} and CPM17~\cite{cpm} datasets by accurately segmenting small and overlapping nuclei and delineating edges precisely. In contrast, other methods tend to merge distinct nuclei into a single entity or over-segment them. These results underscore the exceptional capability of our CC layer in local feature extraction, effectively capturing subtle differences and boundary details between nuclei, thereby significantly enhancing segmentation accuracy.
\begin{figure}[t]
\centerline{\includegraphics[width=\columnwidth]{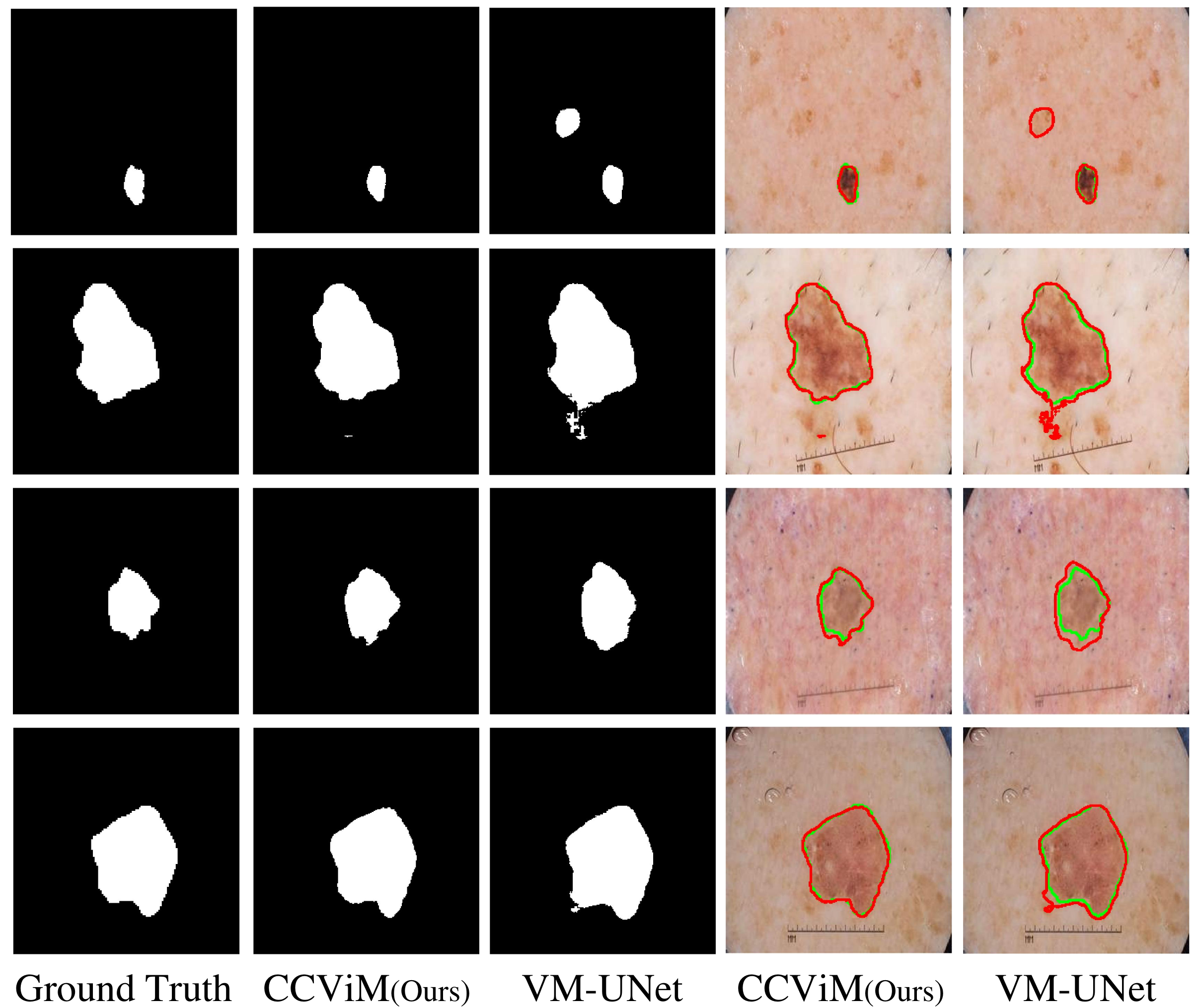}}
\vspace{-2mm}
\caption{Visualizations on the ISIC17~\cite{isic2017} Dataset. The left side presents the ground truth alongside the predicted masks from our model and the VM-UNet~\cite{Vm-unet}. It is evident that our predicted masks are closer to the ground truth. The right side displays the original skin images annotated with lesion contours; green contours denote the ground truth, while red contours indicate the predicted segmentation results. These comparisons further demonstrate the superior effectiveness of our CCViM in accurately segmenting skin lesions.}
\label{fig5}
\vspace{-3mm}
\end{figure}
\begin{figure}[t]
\centerline{\includegraphics[width=\columnwidth]{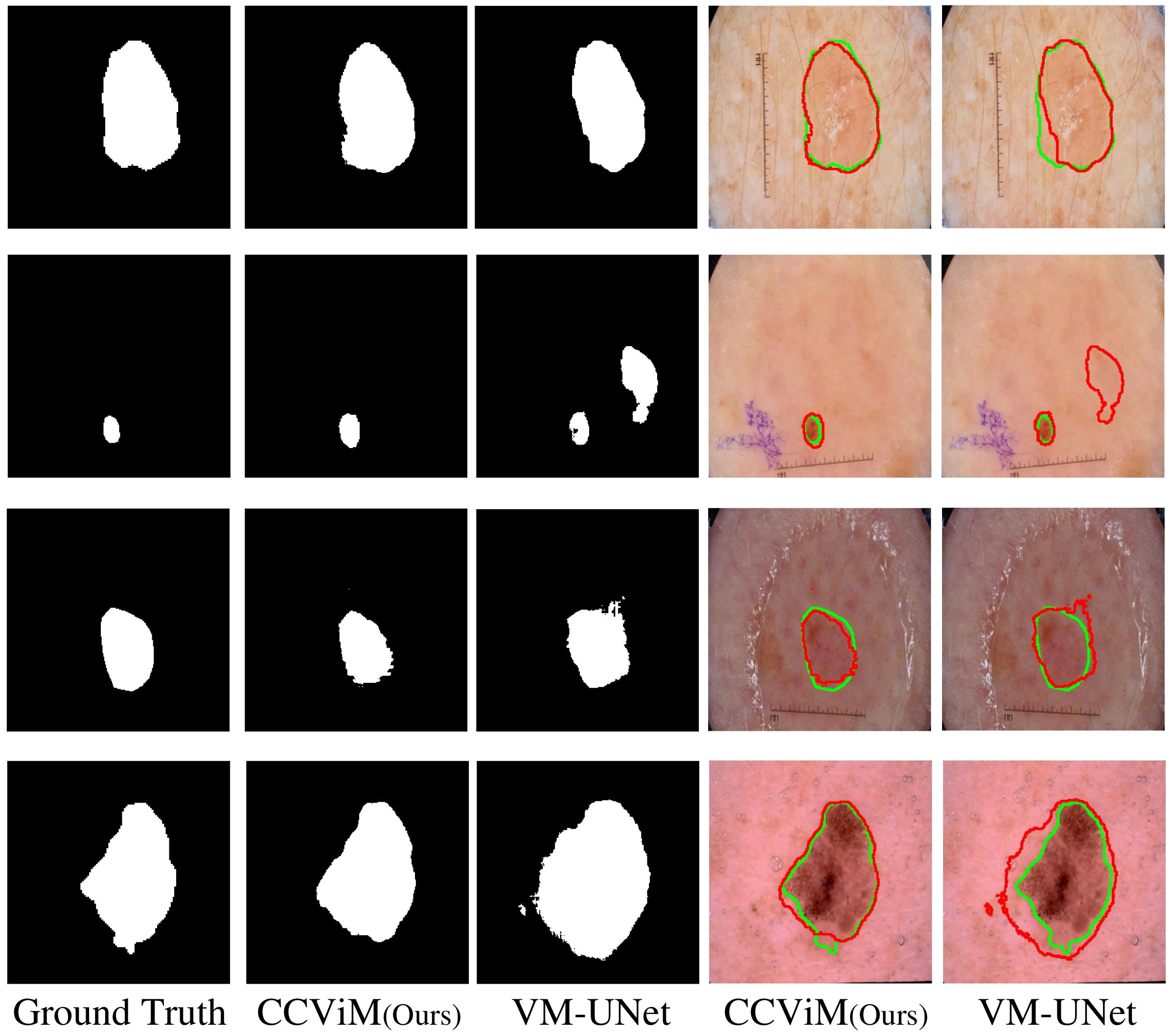}}
\vspace{-2mm}
\caption{Visualizations on the ISIC18~\cite{isic2018} Dataset. The same visualization format as the ISIC17 results described earlier.}
\label{fig6}
\vspace{-4mm}
\end{figure}

\myparagraph{Results on skin image segmentation.}
We compare our CCViM with several state-of-the-art models on the ISIC17~\cite{isic2017} and ISIC18~\cite{isic2018} datasets, Table~\ref{tab2} shows the main results. Compared with CNN-based (\ie, UNet~\cite{unet} and UTNetV2~\cite{UTNetV2}) and Transformer-based (\ie, TransFuse~\cite{Transfuse} and MALUNet~\cite{MALUNet}) methods, the Mamba-based models (\ie, VM-UNet~\cite{Vm-unet}, HC-Mamba~\cite{HC-Mamba}) have the superior performance, which demonstrates the effectiveness of Mamba-based models in MedISeg. From Table~\ref{tab2}, on the ISIC17 dataset, we can observe that our CCViM has improved the mIoU and DSC by $1.17\%$ and $0.71\%$ compared with VM-UNet~\cite{Vm-unet}. On the ISIC18 dataset, our CCViM has improved the mIoU and  DSC by $0.57\%$ and $0.35\%$ compared with VM-UNet~\cite{Vm-unet}. These superior results demonstrate that our CCViM effectively addresses the limitations of Mamba-based models, particularly in preserving local features and spatial context. Fig.~\ref{fig5} shows the visualizations on the ISIC17 dataset, we can observe that our CCViM exhibits accurate and sharp contours, effectively capturing object boundaries. Fig.~\ref{fig6} shows the visualization on the ISIC18~\cite{isic2018} dataset, where our CCViM outputs highlight a closer alignment between the red and green contours, indicating that CCViM is more effective in accurately delineating skin lesions compared to VM-UNet~\cite{Vm-unet}. The VM-UNet~\cite{Vm-unet} often produces over-segmented or merged regions. While VM-UNet~\cite{Vm-unet} performs well in some cases, CCViM demonstrates greater robustness, particularly when handling small lesions and irregular boundaries, offering a closer adherence to the ground truth. These visualizations further demonstrate that the local feature interactions in our CCViM can capture the subtle details of edge contours and irregularly shaped target features.
\begin{table*}[t]
  \centering
  \renewcommand\arraystretch{1.5}
  \setlength{\tabcolsep}{5pt}
  {
   \caption{Result comparisons with state-of-the-art methods on the Synapse~\cite{Synapse} dataset (``-" represents the indicator has not been measured)}
  \begin{tabular}{r|cc|cccccccc} 
    \hline 
     \textbf{Methods} & \textbf{DSC}(\%)\textuparrow & \textbf{HD95}(\%)\textdownarrow & \textbf{Aorta} & \textbf{Gallbladder} & \textbf{Kidney}(\textbf{L}) & \textbf{Kidney}(\textbf{R}) & \textbf{Liver} & \textbf{Pancreas} & \textbf{Spleen} & \textbf{Stomach}
     \\
     \hline
      V-Net~\cite{V-net} & 68.81 & - & 75.34\% & 51.87\% & 77.10\% & 80.75\% & 87.84\% & 40.05\%& 80.56\% & 56.98\%
      \\
      DARR~\cite{DARR} & 69.77 & - & 74.74\% & 53.77\% & 72.31\% & 73.24\% & 94.08\% & 54.18\% & 89.90\% & 45.96\%
      \\
      R50 UNet~\cite{transunet} & 74.68 & 36.87 & 87.47\% & 66.36\% & 80.60\% & 78.19\% & 93.74\% & 56.90\% & 85.87\% & 74.16\%
     \\
      UNet~\cite{unet} & 76.85 & 39.70 & 89.07\% & \cellcolor[gray]{.95}\textbf{69.72}\% & 77.77\% & 68.60\% & 93.43\% & 53.98\% & 86.67\% & 75.58\%
     \\
      R50 Att-UNet~\cite{transunet} & 75.57 & 36.97 & 55.92\% & 63.91\% & 79.20\% & 72.71\% & 93.56\% & 49.37\% &  87.19\% & 74.95\%
     \\
      Att-UNet~\cite{Att-UNet} & 77.77 & 36.02 & 89.55\% & 68.88\% & 77.98\% & 71.11\% & 93.57\% & 58.04\% & 87.30\% & 75.75\%
     \\
     R50 ViT~\cite{transunet} & 71.29 & 32.87 & 73.73\% & 55.13\% & 75.80\% & 72.20\% & 91.51\% & 45.99\% & 81.99\% & 73.95\%
     \\
     TransUNet~\cite{transunet} & 77.48 & 31.69 & 87.23\% & 63.13\% & 81.87\% & 77.02\% & 94.08\% & 55.86\% & 85.08\% & 75.62\%
     \\
     TransNorm~\cite{transnorm} & 78.40 & 30.25 & 86.23\% & 65.10\% & 82.18\% & 78.63\% & 94.22\% & 55.34\% & 89.50\% & 76.01\%
     \\
      Swin U-Net~\cite{Swin-unet} & 79.13 & 21.55 & 85.47\% & 66.53\% & 83.28\% & 79.61\% & 94.29\% & 56.58\% & 90.66\% & 76.60\%
     \\
     TransDeepLab~\cite{transdeeplab} & 80.16 & 21.25 & 86.04\% & 69.16\% & 84.08\% & 79.88\% & 93.53\% & 61.19\% & 89.00\% & 78.40\%
     \\
     UCTransNet~\cite{uctransnet} & 78.23 & 26.75 & - & - & - & - & - & - & - & -
     \\
     MT-UNet~\cite{MT_UNet} & 78.59 & 26.59 & 87.92\% & 64.99\% & 81.47\% & 77.29\% & 93.06\% & 59.46\% & 87.75\% & 76.81\%
     \\
     MEW-UNet~\cite{Mew-unet} & 78.92 & \cellcolor[gray]{.95}\textbf{16.44} & 86.68\% & 65.32\% & 82.87\% & 80.02\% & 93.63\% & 58.36\% & 90.19\% & 74.26\%
     \\
      VM-UNet~\cite{Vm-unet} & 81.08 & 19.21 & 86.40\% & 69.41\% & 86.16\% & 82.76\% & 94.17\% & 58.80\% & 89.51\% & 81.40\%
     \\
      HC-Mamba~\cite{HC-Mamba} & 81.56 & 26.32 & \cellcolor[gray]{.95}\textbf{90.92}\% & 69.65\% & 85.57\% & 79.27\% & 
      \cellcolor[gray]{.95}\textbf{97.38}\% & 54.08\% & 
      \cellcolor[gray]{.95}\textbf{93.49}\% & 80.14\%
     \\
     \cellcolor[gray]{.95}\textbf{CCViM$_{{\textrm{(Ours)}}}$} & \cellcolor[gray]{.95}\textbf{82.65} & 17.83 & 87.63\% & 68.45\% & 
     \cellcolor[gray]{.95}\textbf{86.23}\% & \cellcolor[gray]{.95}\textbf{83.22}\% & 94.67\% & \cellcolor[gray]{.95}\textbf{67.12}\% & 92.05\% & \cellcolor[gray]{.95}\textbf{81.82}\%
     \\
     \hline  
   \end{tabular}
   \label{tab3}
  }
  \vspace{-1mm}
\end{table*}

\myparagraph{Results on synapse multi-organ segmentation.}
We also compare our CCViM with state-of-the-art models on the Synapse dataset, where similar observations and conclusions can be observed. In Table~\ref{tab3}, Mamba-based models have superior performance compared with CNN-based and Transformer-based models. Compared with VM-UNet~\cite{Vm-unet}, our model has improved the DSC by $1.57\%$ and has reduced the HD95 by $1.38\%$. Our model gets state-of-the-art results in segmenting $8$ types of abdominal organs (aorta, gallbladder, left kidney, right kidney, liver, pancreas, spleen, stomach). Fig.~\ref{fig7} shows the visualizations, compared to VM-UNet~\cite{Vm-unet}, we can observe that our CCViM not only segments various organs more accurately but also delineates edges with greater precision. The superior performance further demonstrates the effectiveness of both global and local feature interactions in our CCViM.
\begin{figure}[t]
\centerline{\includegraphics[width=\columnwidth]{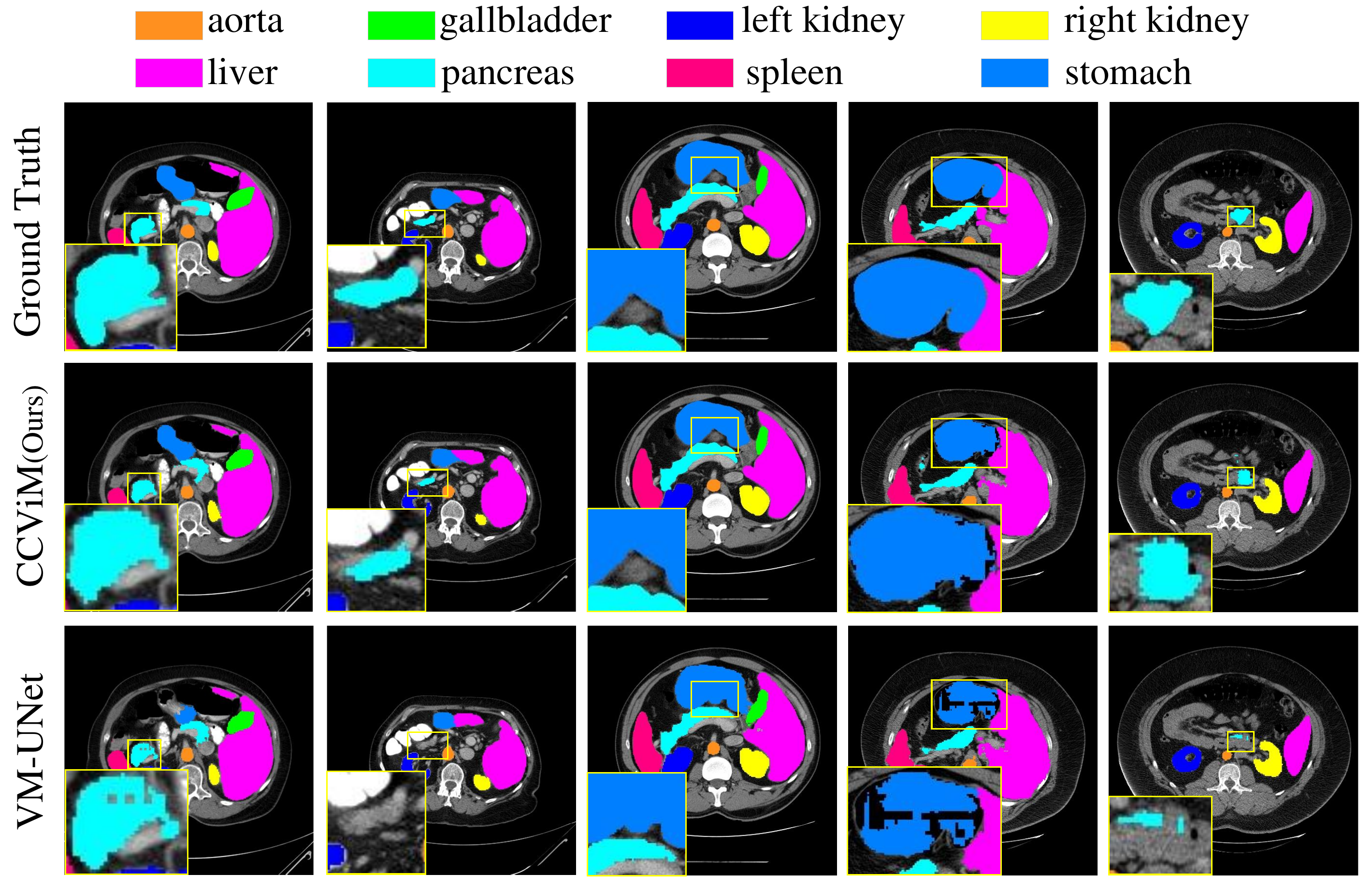}}
\vspace{-3mm}
\caption{Visualizations on the Synapse~\cite{Synapse} Dataset. Compare with VM-UNet~\cite{Vm-unet}, our model is closer to ground truth, which demonstrates our CCViM has superior performance.
}
\label{fig7}
\vspace{-4mm}
\end{figure}
\begin{table}[t]
  \centering
  \renewcommand\arraystretch{1.5}
  \setlength{\tabcolsep}{1pt} 
  {
   \caption{The Superiority of Our CC on the Kumar~\cite{kumar2017dataset} and ISIC17~\cite{isic2017} datasets}
  \begin{tabular}{lr|ccccc} 
    \hline 
     ~ & \textbf{Methods} & \textbf{PQ}(\%)\textuparrow & \textbf{Dice}(\%)\textuparrow & \textbf{AJI}(\%)\textuparrow & \textbf{DQ}(\%)\textuparrow & \textbf{SQ}(\%)\textuparrow 
     \\			
     \hline
     \multirow{4}*{\rotatebox{90}{\textbf{Kumar}~\cite{kumar2017dataset}}} & LocalVIM~\cite{localmamba}  & 49.15 & 79.02 & 53.28 & 67.42 & 72.65
     \\
     ~ & \cellcolor[gray]{.95}LocalVIM-CC & 50.72 & 79.17 & 54.96 & 69.17 & 73.04  
     \\
      ~ & LocalVMamba~\cite{localmamba}  & 58.02 & 82.23 & 61.08 & 75.32 & \cellcolor[gray]{.95}\textbf{76.72}
     \\
     ~ & \cellcolor[gray]{.95}LocalVMamba-CC  &
     \cellcolor[gray]{.95}\textbf{58.51} & \cellcolor[gray]{.95}\textbf{82.49}	& \cellcolor[gray]{.95}\textbf{61.42} & \cellcolor[gray]{.95}\textbf{76.58} & 76.63
     \\
      \hline 
     ~ & \textbf{Methods}& \textbf{mIoU}(\%)\textuparrow & \textbf{DSC}(\%)\textuparrow & \textbf{Acc}(\%)\textuparrow & \textbf{Spe}(\%)\textuparrow & \textbf{Sen}(\%)\textuparrow 
     \\
     \hline
      \multirow{4}*{\rotatebox{90}{\textbf{ISIC17}~\cite{isic2017}}}  & LocalVIM~\cite{localmamba}  & 77.94 & 87.60 & 95.90 & 97.78 & 86.54
     \\
      ~ & \cellcolor[gray]{.95}LocalVIM-CC & 78.23 & 87.78 & 96.04 & 98.24 & 85.08
     \\
      ~ & LocalVMamba~\cite{localmamba}  & 78.70 & 88.08 & 96.07 & \cellcolor[gray]{.95}\textbf{97.97} & 86.62
     \\
     ~ & \cellcolor[gray]{.95}LocalVMamba-CC  & 
     \cellcolor[gray]{.95}\textbf{81.38} & \cellcolor[gray]{.95}\textbf{89.73} & \cellcolor[gray]{.95}\textbf{96.56} & 97.93 & \cellcolor[gray]{.95}\textbf{89.75}
     \\
     \hline  
   \end{tabular}
   \label{tab4}
  }
\vspace{-1mm}
\end{table}
\subsection{Ablation Analysis}
\label{IV-E}
\myparagraph{Superiority of the context clustering (CC) layer.}
To assess the effectiveness of our CC layer, we conduct the ablation experiments on the recent state-of-the-art models LocalVIM~\cite{localmamba} and LocalVMamba~\cite{localmamba} using Kumar and ISIC17 datasets. The local scanning strategies in LocalVIM~\cite{localmamba} and LocalVMamba~\cite{localmamba} can effectively capture local dependencies while maintaining a global perspective, however, the scanning strategies rely on fixed propagation trajectories, which cannot adaptively capture local features and ignore the spatial context information. Our CC layer can cluster the local features and capture spatial context information in an adaptive way. Therefore, we exchange the local scanning strategies in LocalVIM~\cite{localmamba} and LocalVMamba~\cite{localmamba} with our CC layer, and the results are shown in Table~\ref{tab4}. On Kumar, our CC layer has improved the PQ, Dice, AJI, DQ, and SQ metrics by $1.57\%$, $0.15\%$, $1.68\%$, $1.75\%$ and $0.39\%$ respectively, compared to LocalVIM~\cite{localmamba}. And our CC layer has improved the PQ, Dice, AJI, DQ, and SQ metrics by $0.49\%$, $0.26\%$, $0.34\%$, $1.26\%$ and $-0.09\%$ respectively, compared to LocalVMamba~\cite{localmamba}. On the ISIC17 dataset, our CC layer has improved the mIoU and DSC metrics by $0.29\%$ and $0.18\%$ respectively, compared to LocalVIM~\cite{localmamba}. Besides, our CC layer has improved the mIoU and DSC metrics by $2.68\%$ and $ 1.65\%$ respectively, compared to LocalVMamba~\cite{localmamba}. The improvements demonstrate the superiority of our CC layer in dynamically capturing local features and spatial information.

\begin{table}[t]
  \centering
  \renewcommand\arraystretch{1.5}
  \setlength{\tabcolsep}{1pt}
  {
   \caption{The results of different cluster centers on Kumar~\cite{kumar2017dataset} and ISIC17~\cite{isic2017} datasets}
   \vspace{-2mm}
  \begin{tabular}{lr|ccccc} 
    \hline 
    ~ & \textbf{Setting} & \textbf{PQ}(\%)\textuparrow & \textbf{Dice}(\%)\textuparrow & \textbf{AJI}(\%)\textuparrow & \textbf{DQ}(\%)\textuparrow & \textbf{SQ}(\%)\textuparrow 
     \\			
     \hline
     \multirow{13}*{\rotatebox{90}{\textbf{Kumar}~\cite{kumar2017dataset}}} & h-hflip & 56.32 & 81.90 & 60.03 & 74.11 & 75.68
     \\
     ~ & h-hflip-C$_4$ & 57.84 & 81.93 & 59.95 & 75.36 & 76.49
     \\
      ~ &  h-hflip-C$_{25}$ & 57.91 & 82.09 & 60.46 & 75.34 & 76.61
      \\
      ~ &  h-hflip-C$_4$-C$_{25}$ & 58.14 & 82.34 & 60.45 & 75.72 & 76.57
     \\
      ~ &  v-vflip & 57.37 & 81.88 & 59.72 & 74.84 & 76.35
     \\
      ~ &  v-vflip-C$_4$ & 57.06 & 81.92 & 60.09 & 74.89 & 75.95
     \\
      ~ &  v-vflip-C$_{25}$ & 57.84 & 81.97 & 60.63 & 75.68 & 76.14
     \\
     ~ &  v-vflip-C$_4$-C$_{25}$ & 58.49 & 82.26 & 60.86 & 75.96 & 76.73
     \\
     ~ &  h-hflip-v-vflip & 56.59 & 81.89 & 60.00 & 74.27 & 75.85
     \\
      ~ &  h-v-C$_4$-C$_{25}$ & 57.97 & 
      \cellcolor[gray]{.95}\textbf{82.57} & 60.80 & 75.83 & 76.21
     \\
      ~ &  h-vflip-C$_4$-C$_{25}$ & 58.83 & 82.32 & 61.40 & 76.43 & 76.70
     \\
     ~ &  hflip-v-C$_4$-C$_{25}$ & 58.23 & 82.40 & 61.38 & 76.17 & 76.11
     \\
     ~ &  hflip-vflip-C$_4$-C$_{25}$ & 58.49 & 82.37 & 60.95 & 76.21 & 76.44
     \\
     ~ & \cellcolor[gray]{.95}\textbf{CCViM$_{{\textrm{(Ours)}}}$}  & \cellcolor[gray]{.95}\textbf{58.83} & 82.48 & \cellcolor[gray]{.95}\textbf{61.38} & \cellcolor[gray]{.95}\textbf{76.50} & \cellcolor[gray]{.95}\textbf{76.63}
     \\
      \hline 
     ~ & \textbf{Setting}& \textbf{mIoU}(\%)\textuparrow & \textbf{DSC}(\%)\textuparrow & \textbf{Acc}(\%)\textuparrow & \textbf{Spe}(\%)\textuparrow & \textbf{Sen}(\%)\textuparrow 
     \\
     \hline
     \multirow{13}*{\rotatebox{90}{\textbf{ISIC17}~\cite{isic2017}}} & h-hflip & 80.21 & 89.01 & 96.27 & 97.5 & 
     \cellcolor[gray]{.95}\textbf{90.14} 
     \\
      ~ & h-hflip-c4 & 80.91 & 89.45 & 96.49 & 98.03 & 88.83
     \\
      ~ &  h-hflip-C$_{25}$ & 80.86 & 89.42 & 96.46 & 97.92 & 89.2
      \\
      ~ &  h-hflip-C$_4$-C$_{25}$ & 80.36 & 89.11 & 96.41 & 98.16 & 87.74
     \\
      ~ &  v-vflip & 80.38 & 89.12 & 96.42 & 98.23 & 87.44
     \\
      ~ &  v-vflip-C$_4$ & 80.9 & 89.44 & 96.44 & 97.72 & 90.03
     \\
      ~ &  v-vflip-C$_{25}$ & 81.14 & 89.59 & 96.53 & 98.06 & 88.96
     \\
     ~ &  v-vflip-C$_4$-C$_{25}$ & 80.83 & 89.40 & 96.47 & 98.00 & 88.87
     \\
     ~ &  h-hflip-v-vflip & 80.23 & 89.03 & 96.29 & 97.58 & 89.90
     \\
     ~ &  h-v-C$_4$-C$_{25}$ & 80.69 & 89.31 & 96.46 & 98.11 & 88.29
     \\
      ~ &  h-vflip-C$_4$-C$_{25}$ & 80.66 & 89.30 & 96.48 & 98.26 & 87.63
     \\
     ~ &  hflip-v-C$_4$-C$_{25}$ & 81.29 & 89.68 & 96.62 & 98.38 & 87.83
     \\
     ~ &  hflip-vflip-C$_4$-C$_{25}$ & 
     \cellcolor[gray]{.95}\textbf{81.48} & 
      \cellcolor[gray]{.95}\textbf{89.80} & 
      \cellcolor[gray]{.95}\textbf{96.66} & \cellcolor[gray]{.95}\textbf{98.46} & 87.74
     \\
     ~ & \cellcolor[gray]{.95}\textbf{CCViM$_{{\textrm{(Ours)}}}$} & 81.40 & 89.74 &  96.60 & 98.19 & 88.7
     \\
     \hline  
   \end{tabular}
   \label{tab5}
  }
  \vspace{-3mm}
\end{table}
\myparagraph{Hyper-parameter analysis.}
Given the different sizes and numbers of the given images, we adopt two different CC layers. One CC layer method has $4$ cluster centers in a local region, another one has $25$ cluster centers in a local region. Besides, there are four different scan directions from which to choose. Therefore, in this section, we perform comprehensive experiments on Kumar~\cite{kumar2017dataset} and ISIC17~\cite{isic2017} to thoroughly analyze the performance of various scanning directions and CC centers. As shown in Table~\ref{tab5}, the term ``h-hflip" indicates that only apply the horizontal scan and horizontal flipped scan in each CCS6 layer. This configuration, which uses only scanning directions without incorporating CC layers, allows us to evaluate the effect of global scanning alone. The term ``h-hflip-C$_4$" denotes applying the horizontal scan, horizontal flipped scan, and CC with $4$ cluster centers in each CCS6 layer. In this configuration, a CC layer with $4$ cluster centers is added to the global scanning directions. This enables a comparison of how the inclusion of a local CC layer enhances performance. Similarly, the term ``h-hflip-C$_{25}$" represents applying the horizontal scan, horizontal flipped scan, and CC with $25$ cluster centers in each CCS6 layer. This configuration explores the effect of increasing the number of cluster centers in the local CC layer while keeping the global scanning directions fixed. The term ``h-hflip-C$_4$-C$_{25}$" represents applying the horizontal scan, horizontal flipped scan, CC with 4 cluster centers, and CC with 25 cluster centers in each CCS6 layer. By combining both CC layers (with 4 and 25 cluster centers), this configuration allows us to examine the performance of the combined effect of multiple clustering operations. The term ``v-vflip" represents only applying the vertical scan and vertical flipped scan in each CCS6 layer. The following can be analogized. This configuration provides a clearer basis for comparison, allowing us to understand the contributions of global scanning and local clustering without introducing additional variables at each stage. Although further configuration tuning may lead to performance improvements, such adjustments would be redundant and unnecessary. Our focus is on evaluating the effectiveness of different scanning directions and CC layers with varying numbers of cluster centers. Given the complexity of multiple stages, introducing additional configurations would overcomplicate the analysis. On Kumar~\cite{kumar2017dataset}, we observe that incorporating CC with 4 or 25 cluster centers improves results compared to using ``h-hflip" or ``v-vflip" alone. Moreover, combining CC with both 4 and 25 cluster centers yields significant performance improvements. Similarly, on the ISIC17 dataset, the performance has improved after incorporating CC with 4 or 25 cluster centers into each CCS6 layer of the ``h-hflip" or ``v-vflip" configurations. The ``h-hflip-C$_4$-C$_{25}$" configuration yielded relatively poorer results compared to other CC combinations. However, most configurations involving CC combinations achieve satisfactory results. Overall, combining CC with both 4 and 25 cluster centers demonstrated superior performance. Notably, the ``hflip-vflip-C$_4$-C$_{25}$" configuration even outperformed our CCViM on ISIC17~\cite{isic2017}. These superior results demonstrate that the local feature interactions and dynamically capturing spatial information are significant in MedISeg, which also demonstrates our CC is effective.
\begin{table}[t]
  \centering
  \renewcommand\arraystretch{1.5}
  \setlength{\tabcolsep}{2.5pt}
  {
   \caption{Model efffciency comparison regarding parameter number (M) and FLOPs (G) }
   \vspace{-2mm}
  \begin{tabular}{r|ccccc} 
    \hline 
    \textbf{Methods} & \textbf{Params}\textdownarrow & \textbf{FLOPs}\textdownarrow  & \textbf{Thru.}\textuparrow & \textbf{PQ}\textuparrow & \textbf{Dice}\textuparrow
     \\[-2mm]		
     ~ & (M) & (G)  & (fps) & (\%) & (\%)
     \\
     \hline
     UNet~\cite{unet} &  31.04 & 54.75 & 34.86 & 47.80 & 75.80 
     \\
     TransUnet~\cite{transunet} & 91.52 & 29.19 & 36.84 & 48.14 & 77.76 
     \\
     Swin U-Net~\cite{Swin-unet} & 27.15 & 7.73  & 37.51 & 47.46 & 80.11
     \\
     VM-UNet~\cite{Vm-unet} & 
     \cellcolor[gray]{.95}\textbf{22.05} & 
     \cellcolor[gray]{.95}\textbf{4.12}  & \cellcolor[gray]{.95}\textbf{41.54} & 56.59 & 81.89
     \\
     LocalVMamba~\cite{localmamba} & 25.20 & 4.21  & 41.01 & 58.02 & 82.23
     \\
     \cellcolor[gray]{.95}\textbf{CCViM$_{{\textrm{(Ours)}}}$} & 23.56 & 4.45 & 41.16 &
     \cellcolor[gray]{.95}\textbf{58.83 }& 
     \cellcolor[gray]{.95}\textbf{82.48} 
     \\
     \hline  
   \end{tabular}
   \label{tab6}
  }
\end{table}
\begin{figure}[htbp]
\centerline{\includegraphics[width=\columnwidth]{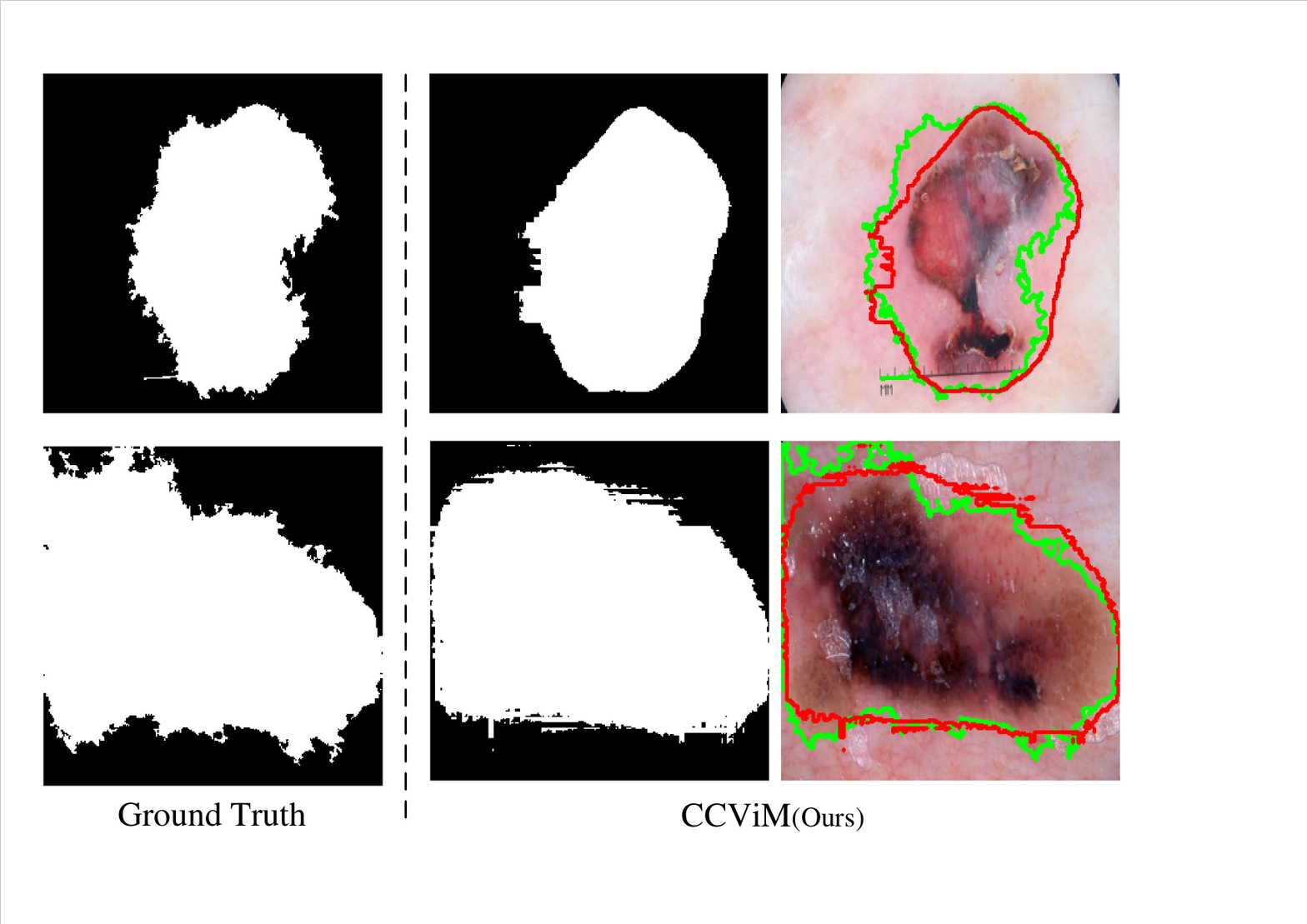}}
\vspace{-3mm}
\caption{Visualizations of some failure examples on ISIC18~\cite{isic2018} dataset.}
\label{fig8}
\vspace{-1mm}
\end{figure}

\myparagraph{Efficiency analysis.} 
We conduct a comparative analysis of our model's parameters and FLOPs against state-of-the-art models, ensuring consistent comparisons by evaluating each model with an input image size of (1,3,256,256). To measure inference speed in a realistic scenario, we use the throughput metric (frames per second, fps), which reflects the average inference speed over the entire test set, calculated across all test images. Additionally, we present results from the Kumar~\cite{kumar2017dataset} dataset, comparing the PQ and Dice metrics. As shown in Table~\ref{tab6}, Mamba-based models, such as VM-UNet~\cite{Vm-unet} and LocalVMamba~\cite{localmamba}, demonstrate superior efficiency compared to CNN-based models (\ie, U-Net~\cite{unet}) and Transformer-based models (\ie, TransUnet~\cite{transunet} and Swin U-Net~\cite{Swin-unet}), offering a balance of lower parameters and FLOPs. Our CCViM model achieves the best performance compared to other methods, as it can extract both global information and capture local spatial features dynamically. Moreover, our CCViM achieves competitive throughput, with an inference speed of 41.16 fps, compared to UNet (\ie, 34.86 fps), TransUnet (\ie, 36.84 fps), Swin U-Net (\ie, 37.51 fps), VM-UNet (\ie, 41.54 fps), and LocalVMamba (\ie, 41.01 fps). The competitive FPS of CCViM can be primarily attributed to the local CC operation. Instead of using traditional convolution or attention mechanisms to extract local features, CCViM uses a context cluster algorithm to aggregate local spatial information dynamically, operating on local windows rather than the entire image. This localized approach significantly reduces computational cost since clustering is performed within small patches rather than over the entire image, which is computationally expensive. Although the CC layer in CCViM introduces a slight increase in FLOPs compared to VM-UNet~\cite{Vm-unet} and LocalVMamba~\cite{localmamba}, this does not notably impact the inference time, allowing CCViM to achieve a throughput that aligns closely with VM-UNet~\cite{Vm-unet} and LocalVMamba~\cite{localmamba},while demonstrating effective segmentation performance.

\subsection{Limitation Analysis}
\label{IV-F}
Although extensive experiments demonstrate the effectiveness of our method in MedISeg tasks, there are two notable limitations. First, while our CC layer may introduce minimal computational costs, this slight addition contributes to promising performance improvements. Second, different medical images may exhibit small lesions and irregular boundaries, which can affect the details captured by our CC layer's local extraction. Additionally, varying scanning directions and CC layers may influence the performance of MedISeg tasks. However, our current configurations for scanning directions and CC layers are fixed, underscoring the need for more adaptive strategies. As shown in Fig. \ref{fig8}, there are several failure examples from the ISIC18~\cite{isic2018} dataset. We can observe that the fixed configuration of global scan directions and local CC layers may impact performance, particularly in capturing the details of irregular boundaries and complex structures. The model struggles to replicate the intricate features represented in the ground truth, leading to discrepancies in segmentation results. An adaptive configuration could enhance our model's ability to discern these features more effectively. Given that our CC layers utilize different clustering centers, varying clustering centers can capture different local details. Furthermore, different scan directions can facilitate varying global interactions, which also affect performance in MedISeg tasks. Adopting a more flexible approach could significantly improve segmentation accuracy by allowing the model to better adapt to the unique characteristics of each medical image. Therefore, developing more adaptive configurations for scan directions and CC layers could enhance performance in both global and local extraction. In the future, we will focus on implementing these adaptive strategies, exploring dynamic configuration algorithms that can adjust in real-time based on the input data characteristics.
\section{Conclusion}
In this paper, we introduce CCViM, a U-shaped architecture designed for medical image segmentation that inherits the efficiency and effectiveness of Mamba. The proposed CC layer partitions feature into distinct windows for learnable local clustering, dynamically capturing spatial contextual information. Based on the CC layer, our CCS6 layer combines the proposed CC with traditional global scanning strategies, significantly enabling our model to capture both local and global information. We compare our CCViM with state-of-the-art models on nuclei segmentation, skin lesion segmentation, and multi-organ segmentation datasets, comprehensive experiments demonstrate the promising performance of our model in medical image segmentation. In the future, we will explore effective searching methods that combine traditional scanning strategies and our CCs in a computation-free way. Furthermore, we will explore the application of Mamba-based methods in diverse medical image recognition tasks, such as registration and reconstruction.

\bibliographystyle{IEEEtran}
\bibliography{ref}
\end{document}